\pdfoutput=1

\documentclass[11pt]{article}

\usepackage{acl}

\usepackage{times}
\usepackage{latexsym}

\usepackage[T1]{fontenc}

\usepackage[utf8]{inputenc}

\usepackage{microtype}
\usepackage{amsmath}
\usepackage{amssymb}
\usepackage{mathtools}
\usepackage{amsthm}
\usepackage{graphicx}
\usepackage{subfigure}
\usepackage{booktabs}
\usepackage{multirow}

\newcommand{\squishlist}{
   \begin{list}{$\bullet$}{%
        \setlength{\itemsep}{0pt}%
        \setlength{\parsep}{0pt}%
        \setlength{\topsep}{-3pt}%
        \setlength{\partopsep}{0pt}%
        \setlength{\listparindent}{-2pt}%
        \setlength{\itemindent}{-5pt}%
        \setlength{\leftmargin}{1em}%
        \setlength{\labelwidth}{0em}%
        \setlength{\labelsep}{0.5em}%
    }
}
\newcommand{\squishend}{
    \end{list}  }

\title{Learning to Maximize Mutual Information for Chain-of-Thought Distillation}

\author{
    Xin Chen$^{1}$, Hanxian Huang$^{2}$, Yanjun Gao$^{3}$, Yi Wang$^{1}$, Jishen Zhao$^{2}$, Ke Ding$^{1}$ \\
    $^{1}$Applied ML Group, Intel Corp. \\
    $^{2}$University of California San Diego, 
    $^{3}$University of Wisconsin Madison \\ 
    \texttt{$^{1}$\{xin.chen, yi.a.wang, ke.ding\}@intel.com,} \\
    \texttt{$^{2}$\{hah008, jzhao\}@ucsd.edu,} \\
    \texttt{$^{3}$ygao@medicine.wisc.edu}
}

\begin{document}
\maketitle
\begin{abstract}
Knowledge distillation, the technique of transferring knowledge from large, complex models to smaller ones, marks a pivotal step towards efficient AI deployment. Distilling Step-by-Step~(DSS), a novel method utilizing chain-of-thought~(CoT) distillation, 
has demonstrated promise by imbuing smaller models with the superior reasoning capabilities of their larger counterparts. 
In DSS, the distilled model acquires the ability to generate rationales and predict labels concurrently through a multi-task learning framework. However, DSS overlooks the intrinsic relationship between the two training tasks, leading to ineffective integration of CoT knowledge with the task of label prediction.
To this end, we investigate the mutual relationship of the two tasks from Information Bottleneck perspective and formulate it as maximizing the mutual information of the representation features of the two tasks. 
We propose a variational approach to solve this optimization problem using a learning-based method. Our experimental results across four datasets demonstrate that our method outperforms the state-of-the-art DSS. 
Our findings offer insightful guidance for future research on language model distillation as well as applications involving CoT. Codes are available at \url{https://github.com/xinchen9/cot_distillation_ACL2024}.  

\end{abstract}

\section{Introduction}
\label{sec:intro}

\begin{figure*}[!ht]
\vskip 0.2in
\begin{center}
\centerline{\includegraphics[width=2\columnwidth]{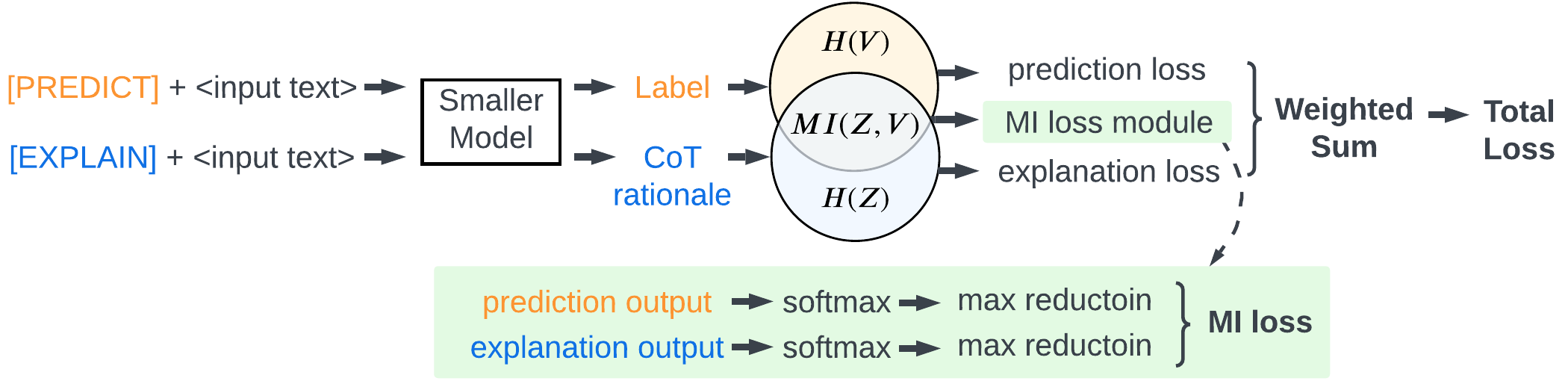}}
\caption{Overview of our approach: CoT distillation from an IB perspective and measurement of the intrinsic relationship between the two tasks by MI. The DSS is an MTL framework pipeline comprising label prediction and rationale generation tasks. 
\textit{H} represents the entropy of representation features \textit{V} and \textit{Z}. Besides prediction loss and explanation losses used in conventional DSS, we design an auxiliary loss module to maximize MI between the two representation features. This process enhances CoT reasoning capacity through knowledge distillation.}
\label{fig:scheme}
\end{center}
\vskip -0.2in
\end{figure*}

The capabilities of larger language models (LLMs) tend to scale with their model size, leading to a substantial demand for memory and compute resources~\cite{chowdhery2023palm,wei2022emergent}. Distilling knowledge from larger LLMs to smaller LLMs has been crucial for the efficient deployment of AI~\cite{Hinton2015distilling,phuong2019towards}. Chain-of-Thought (CoT)~\cite{wei2022chain} distillation represents a pivotal advance in the quest to endow smaller language models with the sophisticated reasoning capabilities of their larger counterparts. By distilling complex thought processes into more compact models, this approach aims to democratize access to advanced natural language understanding and reasoning across a wider array of computational resources~\cite{ma2023sci,magister2023teaching,li2023symbolic}. 

\textit{Distilling Step-by-Step} (DSS)~\cite{hsieh2023distilling} introduces a CoT distillation method that guides smaller models using rationales from LLMs within a multi-task learning (MTL) framework, training them for both label prediction and rationale generation tasks. While DSS brings out the benefits of reducing computational costs, it suffers from the same problem as the conventional MTL framework, that is the difficulty in effectively connecting the prediction and generation tasks. 
The intricacies inherent in training models within the MTL framework can undermine the effectiveness and reliability of the DSS process~\cite{wang-etal-2023-scott}. Despite the successful setup of an MTL framework in DSS, where the tasks of label prediction and rationale generation are intrinsically related, the current configuration may not optimally capture and maximize the mutual knowledge between these tasks. Furthermore, LLMs are prone to producing hallucinations and inconsistent rationales, which 
potentially mislead the student model toward incorrect answers and cause conflicts in MTL that destruct student model learning~\cite{mueller2022text}.


To address this issue, we  
model the DSS using information bottleneck~(IB) and investigate it from an information-theoretic viewpoint~\cite{tishby2015deep}. 
Subsequently, we formulate the DSS as an optimization problem to maximize mutual information~(MI) of label prediction and rationale generation tasks. However, estimating MI from finite data is a known difficult problem~\cite{mcallester2020formal,belghazi2018mutual}. In this study, we introduce a variational method to estimate the MI. Recently, knowledge distillation has been used for solving the challenge of preserving task-specific training in MTL~\cite{li2020knowledge}. Consequently, we propose a practical yet effective auxiliary loss to quantify the shared information between the prediction and the generation tasks, thereby enhancing the alignment between the two tasks and facilitating the knowledge transfer from CoT.  
We conduct comprehensive experiments with two smaller types of T5 models~\cite{2020t5}, T5-base~(220M) and T5-small~(60M), on four popular datasets. Furthermore, we provide detailed analysis in Section~\ref{sec:discussion}. 
Our main contributions are summarized below: 
\squishlist
\item  We reframe the MTL framework of DSS as a MI estimation challenge, aiming to maximize the MI between label prediction and rationale generation tasks. To achieve this, we introduce a variational approach grounded in the IB principle for effective MI estimation. 
To the best of our knowledge, we present the first work of improving CoT distillation from an IB perspective. 
\item Beyond establishing a theoretical foundation, we present a practical approach for MI estimation, incorporating a simple yet effective auxiliary loss to learning to maximize MI and enhance DSS. 
\item  Our methodology demonstrably outperforms existing benchmarks across multiple datasets, evidencing the efficacy of our approach in enhancing the reasoning capabilities of distilled models. 
\item We conduct a systematic review of the relationship between predictive and explainable tasks under MTL training, presenting both qualitative and quantitative analysis results.
\squishend

With the theoretical proofs and experiment results, we aim to provide stepping stones for future research on enhancing CoT distillation through an effective MTL framework, guided by principles from information theory.
\section{Related Work}
\label{sec:related}



We present an overview of previous work across three areas related to our study: knowledge distillation, multi-task learning and information bottleneck. 


\noindent\textbf{Knowledge Distillation~(KD)} While originally designed for training small models by leveraging the extensive knowledge of larger models~\cite{Hinton2015distilling}, KL has been extended to a variety of applications due to its effective transfer of knowledge between models and tasks~\cite{chen2021distilling,wang2021knowledge,sanh2019distilbert,jiao2020tinybert}.
An crucial yet open challenge is how to effectively transfer the knowledge. 
To address the issue, previous studies~\cite{zhang2022contrastive,allen-zhu2023towards,zhang2021self} extract different features and design auxiliary loss functions to enhance KL. 
Our work 
focus on improving the model by acquiring mutual knowledge in addressing both label prediction and rationale generation tasks.

\noindent\textbf{Multi-Task Learning~(MTL)} Through exploiting commonalities and differences of relevant tasks, MTL can enhance improve learning efficiency and prediction accuracy by learning multiple objectives from a share representation~\cite{caruana1997multitask,zhang2021survey}. In recent years, MLT has been broadly applied to  NLP tasks~\cite{worsham2020multi,zhang2023survey,liu-etal-2019-multi}. However, some works figure out that multiple tasks trained simultaneously could conflict with each other and 
it is challenging to optimize the performance of all tasks simultaneously~\cite{kendall2018multi,lin2019pareto}. 
In recent years, KD has also been applied within MTL frameworks, achieving state-of-the-art results in various applications~\cite{li2020knowledge,xu2023multi,yang2022cross}. 

\noindent\textbf{Information Bottleneck ~(IB)} IB~\cite{tishby2015deep,slonim2002information} provides a powerful statistical tool to learn
representation to preserve complex intrinsic correlation
structures over high dimensional data. As a general measure of the
dependence between two random variables, MI is also widely used in deep learning to effectively represent the dependencies of features~\cite{cover1999elements,covert2023learning,liu2009feature}. MI estimation is known to be difficult, and recent progress has been made towards learning-based variational approaches~\cite{tian2019crd,covert2023learning,bachman2019learning,tschannen2019mutual,belghazi2018mutual}. Another challenge associated with the IB principle is the optimization process, which involves a trade-off between achieving a concise representation and maintaining strong predictive capabilities~\cite{alemi2016deep,wang2019deep}. 
Consequently, optimizing IB becomes a complex task that heavily depends on the formulation of the problem and the provision of an effective optimization solution. 
Recent studies has applied IB to solve complex machine learning problems both in computer vision~\cite{tian2021farewell,du2020learning,wan2021multi} and NLP~\cite{chen2020learning,zhang2022improving,paranjape2020information}. In this paper, we formulate our CoT distillation problem with MTL training pipeline using IB and provide a learning-based solution to IB optimization for our CoT distillation problem, as detailed in  Section~\ref{sec:meth}.
\section{Methodology}
\label{sec:meth}


This section begins with an introduction to preliminaries of IB. Following this, we formulate our CoT distillation idea within the IB framework and propose a learning approach to optimize MI.  





\subsection{Preliminaries}
\label{section:pr}

Under the DSS framework, the prefixes \textsc{[predict]} and \textsc{[explain]} will be prepended to the input text, \textsc{text}, for tasks corresponding to label prediction and rationale generation, respectively. In the label prediction task, given the input \textsc{[predict] + text} along with predictive labels $\mathbf{Y}$, a representation feature $\mathbf{V}$, where $\mathbf{V} \in \mathbb{R}^{d}$, is trained using $\mathbf{Y}$. Similarly, in the rationale generation task, the input \textsc{[explain] + text} and rationale label $\mathbf{R}$ guide the training of a representation feature $\mathbf{Z}$, where $\mathbf{Z} \in \mathbb{R}^{d}$, using $\mathbf{R}$.


Our goal is to distill CoT knowledge from larger LLMs to smaller LLMs models. To achieve this, based on the basis of IB~\cite{tishby2015deep,zhang2022improving,wang2019deep}, we model the DSS as following:
\begin{equation}
\label{eq:def}
    I(Z;Y) = \int p(z,y) \log \frac{p(z,y)}{p(z)p(y)}dzdy.
\end{equation}
where sampling observations $z \thicksim \mathbf{Z}$ and $v \thicksim \mathbf{V}$. Here $p(\cdot)$ represents the probability distribution.

To encourage CoT distillation to focus on the information represented in label $\mathbf{Y}$,  we propose using IB to enforce an upper, bound $I_{c}$,  on the information flow from the representation features $V$ to the representation features $Z$. This is achieved by maximizing the following objective:

\begin{equation}
\label{eq:bound}
    \max I(Z;Y)  \ \  s.t. \ \ I(Z;V) \leq I_{c} .
\end{equation}
By employing Lagrangian objective, IB allows $Z$ to be maximally expressive about $Y$ while being maximally compressive regarding the input data, as follows:
\begin{equation}
\label{eq:la}
    \mathcal{C}_{IB} = I(Z;V) - \beta I(Z;Y)  
\end{equation}
where $\beta$ is the Lagrange multiplier. Clearly, Eq.~\ref{eq:la} demonstrates the trade-off optimization between high mutual information and high compression~\cite{zhang2022improving,alemi2016deep}. In our scenario, given a predefined small student model, the compression ratio is fixed. Therefore, we formulate the CoT distillation as an optimization problem:
\begin{equation}
\label{eq:max}
    \max I(Z;V)
\end{equation}
Due to symmetric property of MI, $I(Z;V) = I(V;Z)$. CoT distillation can also enhance rationale generation task with the label knowledge. This is validated in Section ~\ref{sec:discussion}.   
    
\subsection{Variational Bounds of MI}


We rewrite $I(Z;V)$ of Equation~\ref{eq:max} as follows: 
{
\begin{equation}
    \label{eq:mi}
    I(Z;V)  = \mathbb{E}_{p(z,v)}\Big[ \log \frac{p(v|z)}{p(v)} \Big]
\end{equation}
}
According to ~\cite{poole2019variational,covert2023learning}, a tractable variational upper bound can be established by introducing a variational approximation $q(v)$ to replace the intractable marginal $p(v)$, demonstrated by:

\begin{equation} 
\label{eq1}
\begin{split}
I(Z;V) & = \mathbb{E}_{p(z,v)} \Big[\log \frac{p(v|z)q(v)}{p(v)q(v)} \Big] \\
  & = \mathbb{E}_{p(z,v)} \Big[ \log \frac{p(v|z)}{q(v)} \Big] \\
  & - KL(p(v)||q(v))
\end{split}
\end{equation}
here $KL[\cdot||\cdot]$ denotes Kullback-Leibler divergence. The bound is tight when $q(v) = p(v)$. Consequently, $KL(p(v)||q(v))$ is equal to $KL(p(v)||p(v))$, which becomes zero. Therefore, we can derive at the following inequality:

{
\begin{equation}
   I(Z;V) \leq \mathbb{E}_{p(z,v)} \Big[ \log \frac{p(z|v)}{p(v)} \Big]
\end{equation}
}

We can then express the MI from Eq.~\ref{eq:mi} as the follows:
{
\begin{equation}
 \label{eq2}  
 \begin{split}
 \mathbb{E}_{p(z,v)} \Big[ \log \frac{p(z|v)}{p(v)}  \Big]
 &= \sum p(z,v) \log \frac{p(v|z)}{p(v)} \\
 &= \sum p(z|v)p(v)\log p(v|z)  \\
 &- \sum p(v) p(z|v)\log p(v)
 \end{split}
\end{equation}
}
Assuming that  $p(v)$ is uniform distribution for maximal entropy~\cite{schroder2020estimating}, the term $\sum p(v) p(z|v)\log p(v)$ is considered as a constant. This also allows for the omission of $p(v)$ in $\sum p(z|v)p(v)\log p(v|z)$. By combining Eq.~\ref{eq:mi} and Eq.~\ref{eq2}, then maximization of $I(Z;V)$ can be expressed as:

\begin{equation}
 \label{eq3}  
 \begin{split}
 \max I(Z;V) &= \max \mathbb{E}_{p(z,v)} \Big[ \log \frac{p(z|v)}{p(v)}  \Big]\\
 &\varpropto \max  \sum p(z|v) \log p(v|z) \\
 &= \max ( -\sum p(z|v) \log \frac{1}{p(v|z)} )\\
 & =\min ( \sum p(z|v) \log \frac{1}{p(v|z)} ) \\
 &= \min(\sum CE(z|v, v|z))
 \end{split}
\end{equation}
here $CE$ represents cross entropy. Accordingly, CoT distillation in Eq.~\ref{eq:max} is transformed into the problem outlined in the above equation. This problem can be addressed with a learning-based method. To tackle this issue, we have developed a new MI loss that minimizes cross-entropy of representation features of the rationale generation~($p(z|v)$) and representation
features of the label prediction~($p(v|z)$), effectively maximizing MI during CoT distillation process.
\subsection{Training Loss}
\label{sec:loss}
The training loss is given by
\begin{equation}
    \mathcal{L}_{total} 
= \alpha_1 \mathcal{L}_{\mathrm{prediction}} + \alpha_2 \mathcal{L}_{\mathrm{generation}} + \alpha_3 \mathcal{L}_{\mathrm{CE}}
\label{eq: dur_los}
\end{equation}
where $\alpha_1$, $\alpha_2$ and $\alpha_3$ are regularization parameters, all of which are non-negative. $\mathcal{L}_{\mathrm{prediction}}$ represents the loss of the label prediction task, and $\mathcal{L}_{\mathrm{generation}}$ represents the loss of the rationale generation task. Both are general cross-entropy loss as defined in~\cite{hsieh2023distilling}.   

According to the last line of Equation~\ref{eq3}, we define the our MI loss as
\begin{equation}
    \mathcal{L}_{\mathrm{CE}} = l(f(\mathbf{Z}),f(\mathbf{V}))
\end{equation}
$f$ represents our proposed mutual information (MI) loss module, and $l$ denotes the cross-entropy loss. As shown in Figure~\ref{fig:scheme}, the MI loss module consists of softmax and max reduction layers. The softmax function separately calculates the distributions for the outputs of the vocabulary spaces in the label prediction and rationale generation tasks. Subsequently, a max reduction operation is employed to reduce the dimensionality of the predicted outputs from both tasks to a uniform dimension for the loss calculation. Specifically, in the label prediction task, dimensions are reduced from $\mathbb{R}^{m \times d}$ to $\mathbb{R}^{1 \times d}$, and in the rationale generation task, from $\mathbb{R}^{n \times d}$ to $\mathbb{R}^{1 \times d}$.   



\section{Experiments}
\label{sec:exper}
\subsection{Experimental  Setting}
\paragraph{Datasets.} We conducted the experiments on four widely-used benchmark datasets across three different NLP tasks: e-SNLI~\cite{camburu2018snli} and ANLI~\cite{nie2019adversarial} for natural language inference; CQA~\cite{talmor2018commonsenseqa} for commonsense question answering; and SVAMP~\cite{patel2021nlp} for arithmetic math word problems. We used rationale generated by PaLM 540B~\cite{chowdhery2023palm}, which were collected and open-sourced by~\cite{hsieh2023distilling}\footnote{\label{myfootnote}Data and DSS code are from \url{https://github.com/google-research/distilling-step-by-step}.}.

\paragraph{Setup.} Based on CoT properties and the comparative experimental study in~\cite{hsieh2023distilling}, our work adopted T5-base~(220 million) and T5-small~(60 million) to the student models. 
$\alpha_{1}$ and $\alpha_{2}$ were set as $0.5$ and $\alpha_{3}$ is set as $0.1$. We trained our models on one A100 GPU with $80G$ memory. For T5 base model, the training time for the full-size four dataset was approximately 14.4 hours. For T5 small model, the training times was approximately 8.6 hours.

\paragraph{Baselines.} We compare our work with the state-of-the-art DSS~\cite{hsieh2023distilling} by running their open-sourced code and include two other baseline reported in their work: (1) Standard Finetune, which involves using the prevailing pretrain-then-finetune paradigm to finetune a model with ground-truth labels through standard label supervision.~\cite{howard2018universal}. (2) Single-task, which finetunes the model using both of the label and non-CoT rationale as supervision .

\paragraph{Evaluation Settings.} Following the DSS work~\cite{hsieh2023distilling}, we adopt the accuracy as the performance metrics across all four datasets. Higher accuracy indicates better results. Besides accuracy, we also adopt Expected Calibration Errors~(ECE) and Average Confidence Scores to evaluate calibration of the T5-base model. A lower ECE and higher Average Confidence Scores indicate better performance. We adopt GPT-4 to evaluate Quality of CoT examples and subjective analysis. Please refer to our codes for more details.

\subsection{Results}

\noindent\textbf{Experiments of T5-base Model.}We present our experimental results of the T5-base model in Table~\ref{tab:basemodel}. In single-task training, the rationale and label are concatenated into a single sequence, which is treated as the target in training models~\cite{hsieh2023distilling}. Our proposed method consistently achieves better performance than standard fine-tuning and single-task methods on all datasets. Compared to DSS, our method outperforms DSS on ANLI, CQA, and SVAMP, and achieves nearly the same accuracy on e-SNLI

\noindent\textbf{Experiments of T5-small Model.} The experimental results of T5-small model are shown in Table~\ref{tab:smallmodel}. The patterns of the results are similar to those of T5-base. Our proposed method consistently achieves better performance than standard finetuning across all dataset. Compared to DSS, our method outperforms DSS on ANLI, CQA and SVAMP, and is just 0.2\% less accuracy on e-SNLI.

\noindent\textbf{Distillation with LLM Labels.} We conducted an experiment on e-SNLI and ANLI dataset with T5-base model to evaluate the effect of label quality. We distilled the student models using labels generated by 540B PaLM instead of the ground truth. The results are shown in Table~\ref{tab:my_label_llm}. Comparing Table~\ref{tab:basemodel} and Table~\ref{tab:my_label_llm}, we observe the label quality affects the distillation results in both methods. Even With the noisy LLM labels, our model still outperforms DSS on both datasets.

\begin{figure*}[!t]
\vskip 0.2in
\begin{center}
\centerline{\includegraphics[width=2\columnwidth]{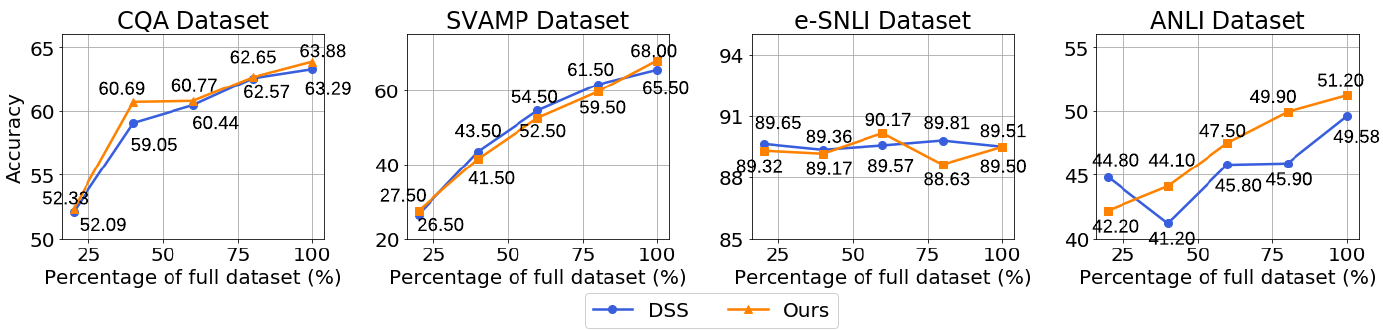}}
\caption{Comparison with DSS with varying sizes of training datasets on T5-base model.}
\label{fig:smalldataset_t5_base}
\end{center}
\vskip -0.2in
\end{figure*}

\begin{figure*}[!t]
\vskip 0.2in
\begin{center}
\centerline{\includegraphics[width=2\columnwidth]{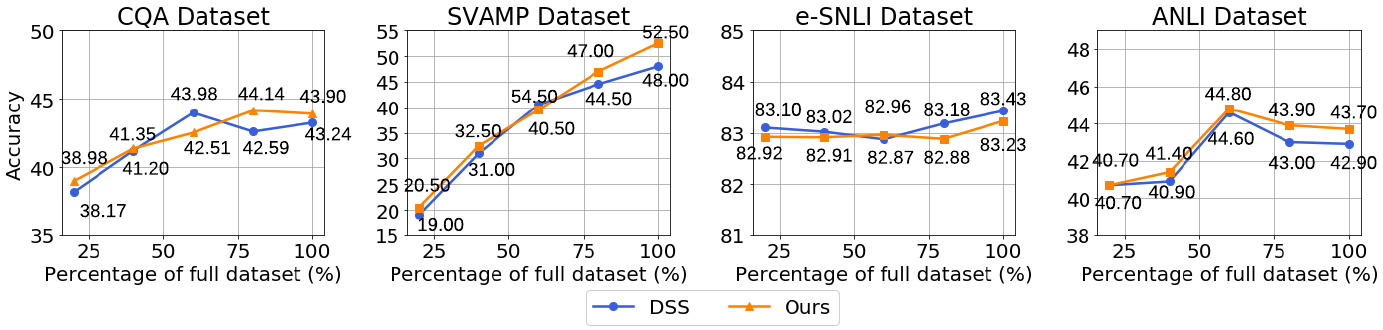}}
\caption{Comparison with DSS with varying sizes of training datasets on T5-small model.}
\label{fig:smalldataset_t5_small}
\end{center}
\vskip -0.2in
\end{figure*}

\noindent\textbf{Distillation with smaller datasets.} To evaluate the performance of our models on smaller datasets, we distilled T5-base and T5-small models on various sizes of four datasets and compared to DSS method. The results are shown in Figure.~\ref{fig:smalldataset_t5_base} and ~\ref{fig:smalldataset_t5_small} respectively.

\begin{table}[!t]
\small
    \centering
    \begin{tabular}{c|c|c|c|c}
    \toprule
        & e-SNLI & ANLI & CQA & SVAMP\\
       \midrule
       Finetuning  & 88.38  & 43.58 & 62.19 &62.63 \\
         \midrule
        Single-task & 88.88 & 43.50 &61.37  & 63.00\\
         \midrule
         DSS & \textbf{89.51} & 49.58 & 63.29 & 65.50\\
         \midrule
         Ours& 89.50 & \textbf{51.20} & \textbf{63.88} & \textbf{68.00}\\
         \bottomrule
    \end{tabular}
    \caption{CoT distillation results on T5-base model. 
    }
    \label{tab:basemodel}
\end{table}
\label{sec:accuracy}


\begin{table}[!tb]
\small
    \centering
    \begin{tabular}{c|c|c|c|c}
    \toprule
        & e-SNLI & ANLI & CQA & SVAMP\\
        \midrule
         Finetuning &  82.90 & 42.00 & 43.16 & 45.00\\
         \midrule
         DSS & \textbf{83.43} & 42.90 & 43.24 & 48.00\\
         \midrule
          Ours & 83.23 & \textbf{43.70} & \textbf{43.90} & \textbf{52.50}\\
         \bottomrule
    \end{tabular}
    \caption{CoT distillation results on T5-small model. 
    }
    \label{tab:smallmodel}
\end{table}

\begin{table}[!tb]
\small
    \centering
    \begin{tabular}{c|c|c} \toprule
      \textbf{Model} &  e-SNLI & ANLI  \\  \midrule
      DSS & 82.65  & 42.80 \\  
       Ours  & 82.81  & 45.50 \\
       \bottomrule
    \end{tabular}
    \caption{Results on two dataset on T5-base model with LLM generated labels.
    }
    \label{tab:my_label_llm}
\end{table}

\subsection{Ablation Study}
\paragraph{Effectiveness of Difference Dimension Reduction Method}
In our proposed MI loss module, we employ maximum reduction to align the dimension of different features. Additionally, mean reduction serves as an alternative method for dimension reduction. We hypothesize that important features can represent better than average features. In Table~\ref{tab:mean_max}, we present the results of two different layer of MI module. The results indicate the superiority of the MI module with maximum reduction.
\paragraph{Comparison with KL Divergence} The KL divergence loss has been extensively utilized in KD tasks, serving as KL a metric for assessing the similarity between two data distributions~\cite{Hinton2015distilling,zhang2022contrastive,gou2021knowledge}. While KL divergence has found widespread application in various KD scenarios, modeling DSS using IB framework proves to be more accurate than using similarity measures, as discussed in Section~\ref{sec:meth}. To validate our hypothesis, we conduct experiments on T5-base model using all four datasets. As shown in Table~\ref{tab:KL_CE}, our proposed method consistently outperforms the KL divergence approach, demonstrating superior performance.  
\begin{table}[!t]
\small 
    \centering
    \begin{tabular}{c|c|c|c|c}
    \toprule 
        & e-SNLI & ANLI & CQA & SVAMP\\
        \midrule 
         Mean &  89.34 & \textbf{51.40} & 63.88 & 66.50\\
         \midrule 
         Max& \textbf{89.50} & 51.20 & \textbf{63.88} & \textbf{68.00}\\
         \bottomrule
    \end{tabular}
    \caption{Results of Mean Reduction Vs Maximum Reduction on T5-based model.  }
    \label{tab:mean_max}
\end{table}

\begin{table}[!ht]
\small 
    \centering
    \begin{tabular}{c|c|c|c|c}
    \toprule
        & e-SNLI & ANLI & CQA & SVAMP\\
        \midrule
         KL Divergence &  89.42 & 42.00 & 62.49 & 67.00\\
         \midrule
         Ours& \textbf{89.50} & \textbf{51.2} & \textbf{63.88} & \textbf{68.00}\\
         \bottomrule 
    \end{tabular}
    \caption{Results of KD loss  VS our proposed cross entropy loss, on T5-base model. }
    \label{tab:KL_CE}
\end{table}


\section{Discussion}
\label{sec:discussion}

\subsection{Analysis on T5 Calibration} 

\begin{table*}[ht]
\centering
\small 
\begin{tabular}{c|cc|cc|cc|cc|cc|cc}
\toprule 
\multirow{2}{*}{\textbf{Model}} & \multicolumn{2}{c|}{\textbf{SVAMP}} & \multicolumn{2}{c|}{\textbf{CQA}} & \multicolumn{2}{c|}{\textbf{e-SNLI}} & \multicolumn{2}{c|}{\textbf{ANLI}} & \multicolumn{2}{c|}{\textbf{e-SNLI} (Out)} & \multicolumn{2}{c}{\textbf{ANLI} (Out)} \\ 
                       & ECE        & Conf.       & ECE       & Conf.      & ECE         & Conf.        & ECE        & Conf.       & ECE                   & Conf.                  & ECE                   & Conf.                 \\ \midrule 
DSS  & 11.81 & 32.56 & 11.75 & 42.79 & 8.54 & 34.33 & 11.12 & 42.72 & 9.81 & 38.01 & 12.78 & 41.69\\
Ours  & 18.92 & 36.81 & 13.65 & 41.17 & 4.35 & 30.06 & 6.94 & 35.90 & 6.61 & 38.08 & 12.27 & 42.35\\
\bottomrule 
\end{tabular}
\caption{Comparisons of our model and DSS on the expected calibration errors (ECE) and average confidence scores (Conf.). }
\label{tab:calib}
\end{table*}
Calibration measures the alignment between a model's predicted accuracy and its confidence levels. \citeauthor{lee2022hard} (\citeyear{lee2022hard}) introduced an innovative perspective on model distillation, positioning the teacher model not only as a source of knowledge but also as a tool for identifying mis-calibration during the training of the student model. This ability to maintain calibration and make reliable predictions is crucial for downstream applications and has been the focus of prior studies~\cite{chen-etal-2023-close, lee2022hard, jiang2021can}. Here, we apply the Expected Calibration Errors (ECE) and Average Confidence Scores to reflect the alignment between the model's predicted probabilities and the actual outcomes, thereby gauging the reliability and certainty of its predictions. Despite the potential limitations inherent in these metrics, we still employ ECE in our experiments due to its simplicity and popularity, as in previous work on investigating the calibration quality of T5~\cite{chen-etal-2023-close, lee2022hard}.  

We employ a 10-bin-based ECE metric and a softmax-based approach to compute average confidence scores from the test outputs across all four datasets. Given that e-SNLI and ANLI essentially represent the same task, we conduct an out-of-domain experiment by testing the model checkpoint trained on one dataset with the test set of the other. This analysis gives us insights into how well our model generalizes across similar tasks and the robustness of its predictions in out-of-domain scenarios and to assess the calibration quality of the model more comprehensively.  

Table~\ref{tab:calib} presents the results of the distilled model calibration evaluation. Overall, both models report lower ECE and confidence scores on SVAMP and e-SNLI, indicating that these two tasks are more challenging and models are less certain about their prediction. Lower ECE values from our MI-based distillation approach are presented for e-SNLI and ANLI, and their respective out-of-domain tests. Notably, our method achieves an ECE of 4.35 in e-SNLI, significantly lower than DSS's 8.54. However, in SVAMP and CQA, our method records higher ECE, indicating potential areas for improvement in these domains. The trade-off in calibration accuracy in specific tasks like SVAMP and CQA compared to DSS suggests future directions for refining our approach. 

Regarding average confidence scores (Conf.), our method generally maintains competitive confidence levels, with notable improvements in e-SNLI and ANLI. In e-SNLI, the confidence is lower (30.06) compared to DSS (34.33), which, combined with a lower ECE, suggests a more realistic confidence estimation. Conversely, in the out-of-domain scenarios for e-SNLI and ANLI, our method shows marginally higher confidence scores than DSS, which, coupled with the lower ECE, indicates robustness in out-of-domain generalization. 


\subsection{Analysis on CoT Output}

\subsubsection{Quality of CoT Examples by GPT-4 Evaluation }
We evaluate the quality of CoT examples using GPT-4, as it achieves the state-of-the-art human alignment performance and is used for text generation evaluation in previous work~\cite{liu2023gpteval,hsu2023gpt,wang-etal-2023-towards}. Inspired by~\cite{wang-etal-2023-towards}, we ask GPT-4 to evaluate the quality of the provided CoT examples based on their coherency and relevancy to the input questions and answers. We randomly sample 50 CoT examples from the four datasets and ask GPT-4 to score based on a scale from 1 to 5, where 1 indicates completely incoherent and irrelevant responses, and 5 represents highly coherent, relevant, and helpful responses. For each sample, we run the same sample for four times to obtain self-consistency to measure the reliability of the responses. Table~\ref{tab:gpt_eval} presents the prompt we use for GPT-4 evaluation,  average scores and standard deviation on the scores obtained over the four datasets. We report the scores on both provided CoT (``gold'') rationales and distilled model predicted rationales. 

\begin{table}[!ht]
\small
\centering
\begin{tabular}{c c c c c}
\toprule
\multicolumn{5}{c}{\textbf{Prompt for GPT 4 Evaluation}} \\ \midrule  
\multicolumn{5}{l}{Given an input pair of a question and an answer of a}\\
\multicolumn{5}{l}{$task name$  task, how good is the given Chain-of-thought} \\
\multicolumn{5}{l}{example? From 1-5, where 1 is completely incoherent and }\\
\multicolumn{5}{l}{irrelevant, 2 is somewhat incoherent and irrelevant, 3 is } \\
\multicolumn{5}{l}{coherent, relevant but not helpful, 4 is somewhat helpful, } \\ 
\multicolumn{5}{l}{and 5 is helpful and it explains the answer well.} \\ 
\midrule
\multicolumn{5}{c}{\textbf{Average Scores and Standard Deviation}} \\ \midrule 
 \textbf{Model}& \textbf{SVAMP} & \textbf{CQA} & \textbf{e-SNLI} & \textbf{ANLI} \\ \midrule 
  Gold &   4.63\scriptsize{$\pm$1.05}  &    3.95\scriptsize{$\pm$1.16}   &   2.42\scriptsize{$\pm$1.23}   &  3.82\scriptsize{$\pm$1.26}  \\ 
  \multicolumn{1}{r}{\scriptsize{\textit{++}}} & 4.43\scriptsize{$\pm$1.18} & 4.11\scriptsize{$\pm$1.40} & 3.49\scriptsize{$\pm$1.35} & 4.01\scriptsize{$\pm$1.10} \\ 
DSS & 2.50\scriptsize{$\pm$1.42}  &    3.60\scriptsize{$\pm$1.61}     &    3.24\scriptsize{$\pm$1.27}   &   3.48\scriptsize{$\pm$1.40}  \\ 
\multicolumn{1}{r}{\scriptsize{\textit{++}}} & 2.53\scriptsize{$\pm$1.46}  &    3.64\scriptsize{$\pm$1.62}     &    3.18\scriptsize{$\pm$1.21}   &   3.44\scriptsize{$\pm$1.30}  \\ 

   Ours & 2.30\scriptsize{$\pm$1.54}  &    3.70\scriptsize{$\pm$1.45}     &    3.03\scriptsize{$\pm$1.47}   &   3.42\scriptsize{$\pm$1.37}  \\ 
   \multicolumn{1}{r}{\scriptsize{\textit{++}}} &  2.72\scriptsize{$\pm$1.45} & 3.63\scriptsize{$\pm$1.60} & 3.17\scriptsize{$\pm$1.17} & 3.34\scriptsize{$\pm$1.21}  \\ 
\bottomrule
\end{tabular}
\caption{Prompt used and results of 50 randomly sampled CoT examples from the four datasets evaluated by GPT-4. We use \textit{++} to denote the setting with \textit{self-consistency} evaluation. }
\label{tab:gpt_eval}
\end{table}

\begin{table}[ht]
\small 
    \centering
    \begin{tabular}{c|c|c|c|c} \toprule
      \textbf{Model} &  \textbf{SVAMP} & \textbf{CQA} & \textbf{e-SNLI} & \textbf{ANLI}  \\  \midrule
      DSS & 0.12 & \textcolor{teal}{0.66} & 0.05  & 0.26 \\ 
      & \scriptsize{$p>0.05$} & \scriptsize{$p<0.05$} & \scriptsize{$p>0.05$} & \scriptsize{$p>0.05$} \\ 
       Ours  &  \textcolor{teal}{0.42}  & \textcolor{teal}{0.53} &  0.03 & 0.26 \\
       & \scriptsize{$p<0.05$} & \scriptsize{$p<0.05$} & \scriptsize{$p>0.05$} & \scriptsize{$p>0.05$} \\  
       \bottomrule
    \end{tabular}
    \caption{ Pearson correlation between CoT quality and accuracy of label prediction on the 50 random samples on the test set. We highlight the correlation with statistical significance ($p<0.05$). }
    \label{tab:my_label}
\end{table}

The effectiveness of our MI-based distillation method is closely linked to the quality of CoT reasoning in the training data. When the CoT quality is high, as in SVAMP, a strong correlation is observed between the model's label prediction accuracy and the quality of its generated CoT. However, this correlation weakens significantly when the CoT quality is low (e-SNLI), suggesting that the model struggles to align label prediction with coherent rationale generation under poor training conditions. Interestingly, with average-quality CoT data (ANLI), the performance gap between our MI-based distillation and DSS is minimal, suggesting that the effectiveness of our approach is particularly reliant on the presence of high-quality reasoning in the training data.

\subsubsection{Case Studies on the Output Rationale}

\begin{figure}[!t]
    \centering
    \includegraphics[width=.5\textwidth]{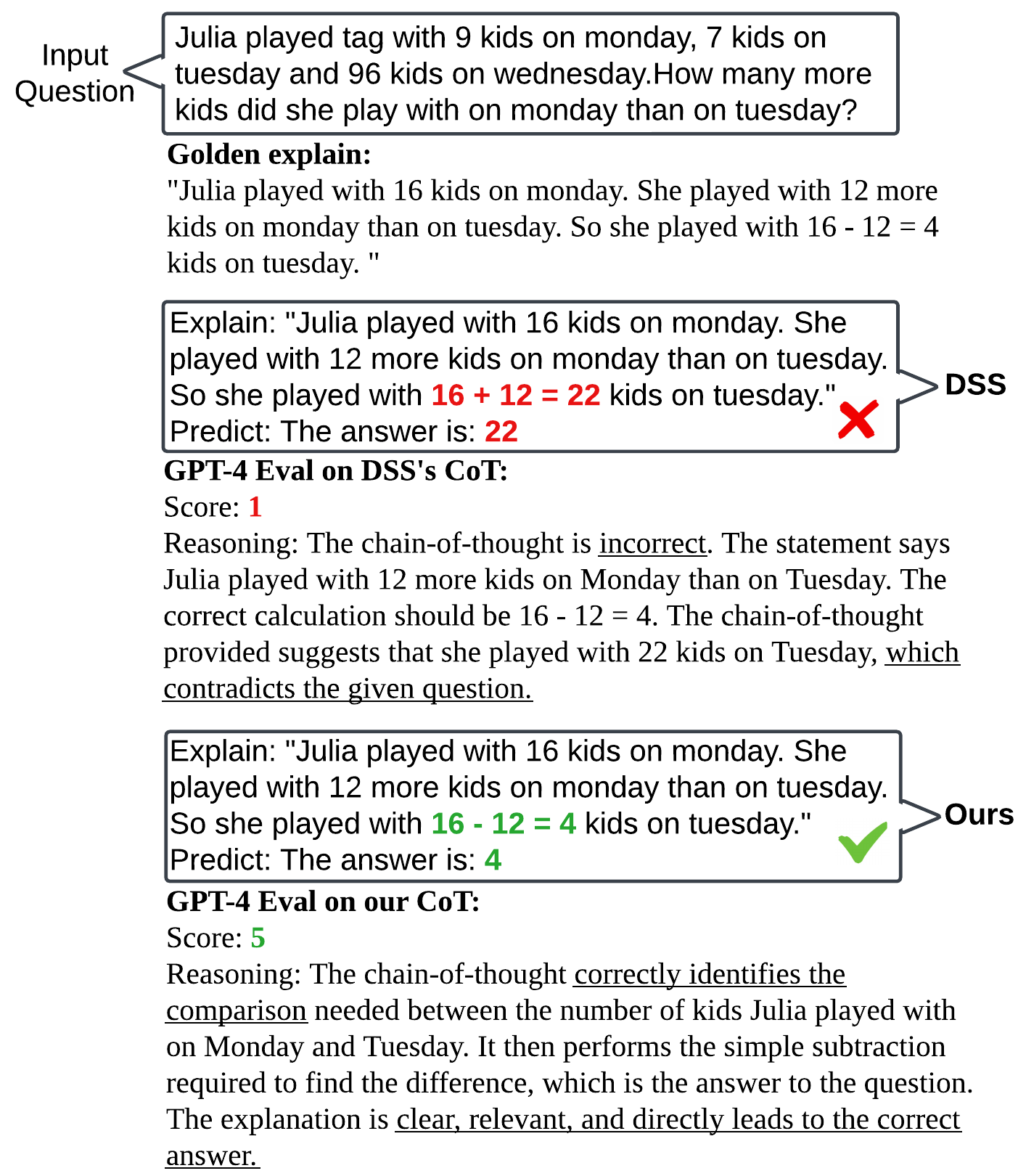}
    \vspace{-10pt}
    \caption{A case study of the output rationale on SVAMP.}
    \label{fig:svamp_example}
\end{figure}

\begin{figure}[!t]
    \centering
    \includegraphics[width=.5\textwidth]{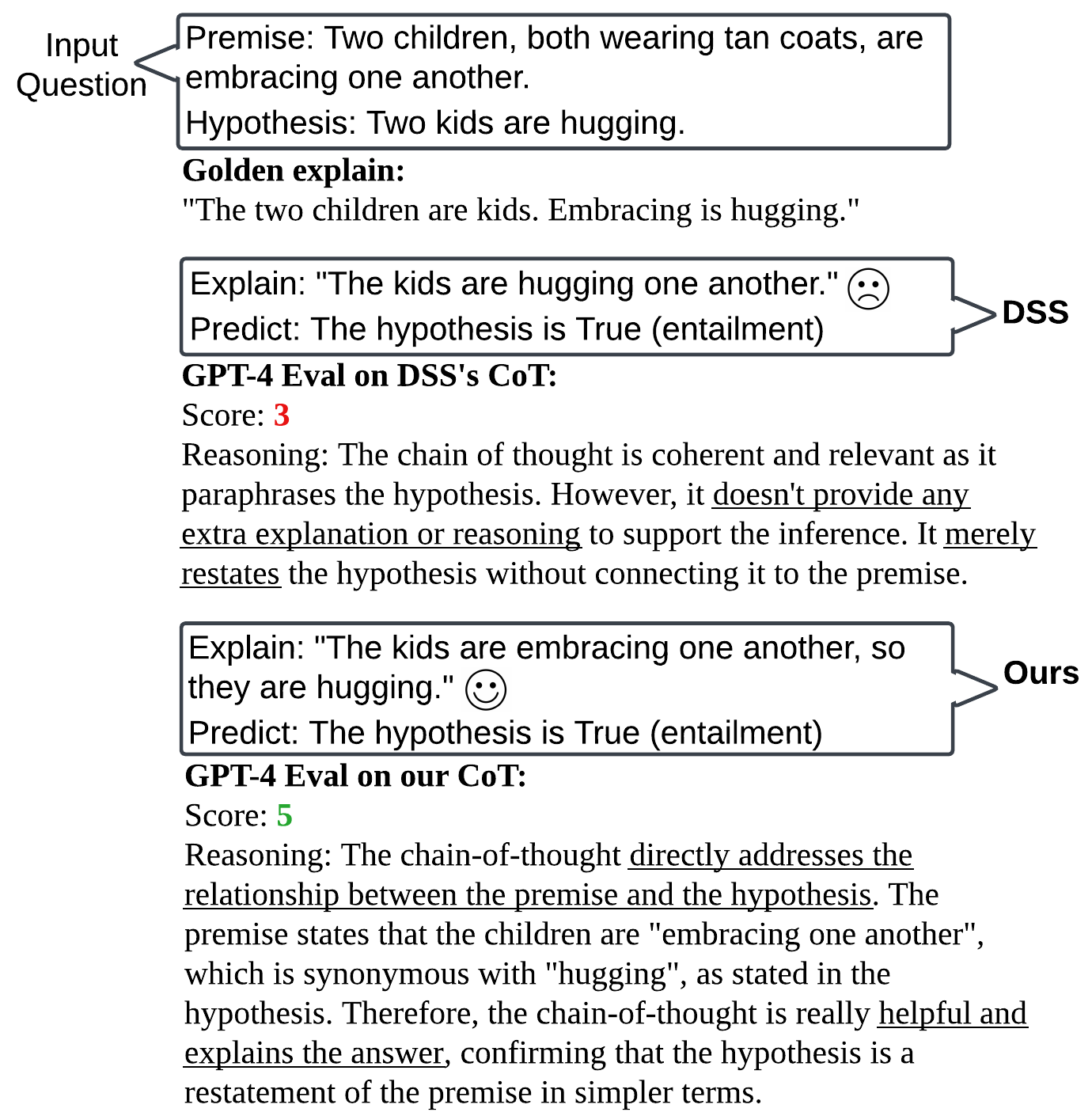}
    \vspace{-10pt}
    \caption{A case study of the output rationale on e-SNLI. 
    }
    \label{fig:esnli_example}
\end{figure}

We performed case studies on SVAMP and e-SNLI as illustrated in Figure~\ref{fig:svamp_example} and~\ref{fig:esnli_example}. In the SVAMP example, the question asks the difference in the number of kids Julia played with from Monday to Tuesday, with specific numbers provided for Monday, Tuesday, and Wednesday. 
DSS generates an incorrect explanation, which contradicts the given question, resulting in to a wrong answer. 
Conversely, our method correctly identifies the
comparison needed between the number of kids Julia played with on Monday and Tuesday, leading to the correct answer. 
Notably, our generated CoT reasoning is identical to the golden one, demonstrating that by precisely grasping the rationale, our approach effectively resolves the math problem. 
We also show the evaluation results (score and reasoning) from GPT-4, where our method gains a top score of 5 and DSS gains only a mere score of 1. 
This example showcases that the high-quality CoT generated by our method enhances problem-solving capabilities in math tasks like SVAMP.

Another example (Figure~\ref{fig:esnli_example}) is from e-SNLI, where the task is to identify whether the hypothesis is entailment, contradiction, or neutral, based on the given premise and hypothesis. Although both our method and DSS generate the correct label output, it is worth noting that, the CoT of our method points out the relationship between the premise and the hypothesis, while DSS only restates the hypothesis without providing any extra explanation or connecting the hypothesis to the premise. Our generated rationale also gains a higher score than DSS. A higher-quality rationale tends to facilitate more accurate label prediction, thereby enhancing overall task performance.
\section{Conclusion}
\label{sec:con}
In this paper, we re-investigate the DSS framework from a information-theoretic perspective. We model it using Information Bottleneck and propose to strengthen it by maximizing the mutual information between rationale generation and label prediction tasks. The proposed learning-based method can automatically optimize the CoT distillation and bolster the reasoning ability of the distilled small models. Our qualitative and quantitative analysis demonstrate the rationale behind our method and shed light on language model distillation and CoT applications.

\section{Limitation}
\label{sec:lim}
Our comparative analysis primarily focuses on the Distilling Step-by-Step (DSS) framework, which serves as our main benchmark. This concentrated comparison, while valuable for a deep understanding of DSS's nuances and our advancements over it, constitutes a limitation of our work. Specifically, our analysis does not extend to a broader range of knowledge distillation methods currently employed in the field. This focus may overlook the potential insights and contrasts that could emerge from evaluating our approach against a wider array of distillation techniques. Future research could benefit from a more expansive comparative study, incorporating diverse methodologies to fully contextualize our findings within the broader landscape of knowledge distillation practices. This broader comparison would not only validate the efficacy of our method in various settings but also illuminate areas for further refinement and innovation.

However, it is important to note that our contribution lies in providing an in-depth analysis from both theoretical and practical viewpoints to enhance the CoT distillation process. Our work delves into the intricacies of utilizing mutual information to improve distillation outcomes, offering significant advancements in understanding and applying CoT distillation techniques. 

\section{Ethical Issues}
\label{sec:ethical}

In this paper, we carefully considered the ethical implications in line with the ACL code of ethics. We evaluated the potential dual-use concerns, ensuring our research serves to benefit society and does not cause inadvertent harm. Our methodology and applications were thoroughly assessed for fairness, non-discrimination, and privacy, particularly in the context of data handling and model outputs. We also ensured our study did not expose any negative impact on individuals and groups. Moreover, we did not engage in academic dishonesty and adhered to high-quality processes and product standards in our professional work. We include this detailed discussion of these ethical considerations, affirming our commitment to responsible and beneficial computational linguistics research.



\bibliography{example_paper,distill,gpt,resource,prompt,multitask,information}

\bibliographystyle{acl_natbib}

\end{document}